\documentclass[12pt]{article}
\usepackage{graphicx}
\usepackage[numbers, sort&compress]{natbib}
\usepackage{xcolor}
\usepackage{multirow}
\usepackage{array}
\usepackage{url} 
\usepackage[hidelinks]{hyperref}
\usepackage{multirow}
\usepackage{booktabs} 
\usepackage{amsmath}

\title{Automatic Classifiers Underdetect Emotions Expressed by Men}
\author{Ivan Smirnov$^{1,5\ast}$, Segun T. Aroyehun $^{2}$, Paul Plener$^{3,4}$, David Garcia$^{2,5}$
\\
\\
\normalsize{$^{1}$University of Technology Sydney, Australia}\\
\normalsize{$^{2}$University of Konstanz, Germany}\\
\normalsize{$^{3}$Medical University of Vienna, Austria}\\
\normalsize{$^{4}$University of Ulm, Germany}\\
\normalsize{$^{5}$Complexity Science Hub, Austria}\\
\\
\normalsize{$^\ast$ivan.smirnov@uts.edu.au}
}

\date{}

\begin{document}
\baselineskip20pt
\maketitle

\begin{abstract}
\normalsize
\baselineskip18pt
The widespread adoption of automatic sentiment and emotion classifiers makes it important to ensure that these tools perform reliably across different populations. Yet their reliability is typically assessed using benchmarks that rely on third-party annotators rather than the individuals experiencing the emotions themselves, potentially concealing systematic biases. In this paper, we use a unique, large-scale dataset of more than one million self-annotated posts and a pre-registered research design to investigate gender biases in emotion detection across 414 combinations of models and emotion-related classes. We find that across different types of automatic classifiers and various underlying emotions, error rates are consistently higher for texts authored by men compared to those authored by women. We quantify how this bias could affect results in downstream applications and show that current machine learning tools, including large language models, should be applied with caution when the gender composition of a sample is not known or variable. Our findings demonstrate that sentiment analysis is not yet a solved problem, especially in ensuring equitable model behaviour across demographic groups.
\end{abstract}

\section*{Introduction}
Automatic sentiment and emotion classifiers have become ubiquitous in both industry and research applications \citep{liu2022sentiment}. Traditionally, their reliability is attributed to the relatively high accuracy rates achieved in standardised benchmarks such as TweetEval \citep{barbieri-etal-2020-tweeteval}. However, these benchmarks almost always depend on datasets in which sentiment or emotions have been labelled by third-party annotators, rather than the individuals experiencing these emotions themselves \citep{rosenthal-etal-2017-semeval}. If the annotators are systematically mislabelling sentiment and emotions, then the benchmarks could overestimate the true accuracy and hide potential biases of the automatic classifiers. If such systematic biases do exist, prior research provides competing evidence about their potential direction, allowing us to formulate two alternative hypotheses.

\medskip

\textbf{H1a: Classification error is higher for women}.

Machine learning algorithms often underperform when applied to women's data \citep{tatman2017gender, buolamwini2018gender, stanovsky2019evaluating, sobhani2024towards}. This is likely explained by the underrepresentation of women in training datasets. For example, men are cited three times more often than women in media datasets \citep{asr2021gender}, and women appear three times less frequently than men in commonly used image captioning datasets \citep{tang2021mitigating}. Additionally, annotators may hold stereotypes about gendered language use \citep{nemani2024gender}, or there may be mismatches between authors' and annotators' linguistic and social norms \citep{hovy2021five}. This could disadvantage women authors when annotators are predominantly men.

\medskip

\textbf{H1b: Classification error is higher for men}.

Women demonstrate greater emotional expressivity, particularly for positive emotions \citep{chaplin2015gender}, while men consistently score higher on measures of difficulty in describing and communicating feelings to others \citep{mendia2024gender}. This might make it harder for automatic classifiers to accurately detect emotions in men's text. Indeed, it was shown that it was easier to predict ratings on TripAdvisor from women reviews \citep{thelwall2018gender}, and the accuracy of speech emotional classification was higher for women \citep{lin2024emo}.

\medskip

To test which of these hypotheses is more strongly supported, we used a unique dataset of social media posts that are self-annotated both for gender and the emotional state of the author. This data comes from TalkLife\footnote{\url{https://www.talklife.com/}}, a social media platform specifically designed for users to share their emotional experiences. Each post on the platform is tagged with a mood label that authors use to communicate their emotional state. The dataset comprises $6,633,562$ posts made by $316,387$ users (see \nameref{sec:methods} for details).

We focus on $16$ mood tags and map them to five broader categories (`Sadness', `Anger', `Fear', `Affection', `Happiness') based on dimensional models of emotion from affective science literature \citep{aroyehun2023leia}. The list of selected mood tags and their corresponding mappings to emotion and sentiment labels is presented in \autoref{tb:map}.

Using a pre-registered design\footnote{\url{doi.org/10.17605/OSF.IO/8FMYT}}, we analyse whether sentiment classification and emotion detection models result in different error rates when applied to posts written by men and those written by women. We compare commonly used machine learning classifiers, most popular dictionary based methods, as well as large language models (LLMs) (see \autoref{tb:model_list} for the full list of models). Since different models use different output labels (e.g., NRC includes label `trust' while other models do not), we standardise all labels to one of three sentiment categories: \textit{positive}, \textit{negative}, or \textit{neutral} (see \autoref{tb:mapping} for mapping that we use). We then compute two error values for each mood tag: the \textit{valence error} as the proportion of cases where classifiers return positive labels instead of negative self-labels, or vice versa; and the \textit{salience error} as the proportion of neutral labels returned by classifiers instead of negative or positive self-labels. Note that \textit{salience error} is computed only for those classifiers that can return a \textit{neutral} label.
After identifying a bias, we estimate how it might affect conclusions drawn from applying automatic classifiers to gender-heterogeneous datasets.

Prior research examining how demographic attributes, such as race or gender, influence sentiment and emotion predictions has often relied on synthetic data or inferred demographic labels. For example, the Equity Evaluation Corpus (EEC) \citep{kiritchenko-mohammad-2018-examining} is a benchmark dataset designed to measure bias in sentiment analysis systems. This approach relied on synthetic sentence templates that varied only by the presence of specific gendered or racial identifiers. Other work \citep{ODBAL2022422} extended the focus to gender-specific disparities in emotion detection models, analysing how training data imbalances and feature associations derived from datasets with inferred demographics contribute to biased performance across gender groups, and proposing mitigation strategies to reduce such biases during model training.

While these studies examine bias using synthetic text or inferred attributes, our work takes a different approach by analysing emotion expressions from real users who self-report their demographic characteristics. This allows us to assess bias in how models interpret genuine emotional language across demographic groups, rather than under synthetic text conditions. Furthermore, we rely on self-reported emotion labels rather than labels assigned by external annotators. We argue that emotions are inherently subjective and best described by the person experiencing them in their natural environment. To the best of our knowledge, only very few datasets have systematically captured such first-person affective experiences at scale \citep{lykousas2019sharing}. Taken together, this approach provides a complementary and ecologically valid perspective on bias in emotion classification.

Moreover, prior approaches are largely limited to examining aggregate performance differences, focusing on overall accuracy rather than developing an understanding of specific error patterns, such as mislabelling across valence categories or systematic underdetection. In contrast, we first analyse affective language at multiple levels of granularity, including sentiment, discrete emotions, and mood tags, to capture how different bias patterns emerge across these affective dimensions. Second, to move beyond overall performance, our study investigates error patterns by introducing two metrics that quantify distinct types of errors: valence error for mislabelling across valence categories and salience error for the underdetection of emotion categories.

Our evaluation spans a range of model types, from dictionary-based approaches to large language models, allowing for a comparative assessment of how bias manifests across different modelling paradigms. This comprehensive approach provides a methodological framework for examining fairness in affect-related text analysis.

Our findings call for a critical reexamination of sentiment and emotion detection through the lens of fairness, demonstrating that sentiment analysis remains far from a solved problem, especially in ensuring equitable model behaviour across demographic groups. By combining ecologically valid data with a systematic error evaluation framework, our work provides a foundation for future research aimed at building responsible affective computing systems that more consistently and accurately reflect emotional expression across diverse user populations.

\section*{Results}
\begin{figure}[h!]
\centering
\includegraphics[width=14cm]{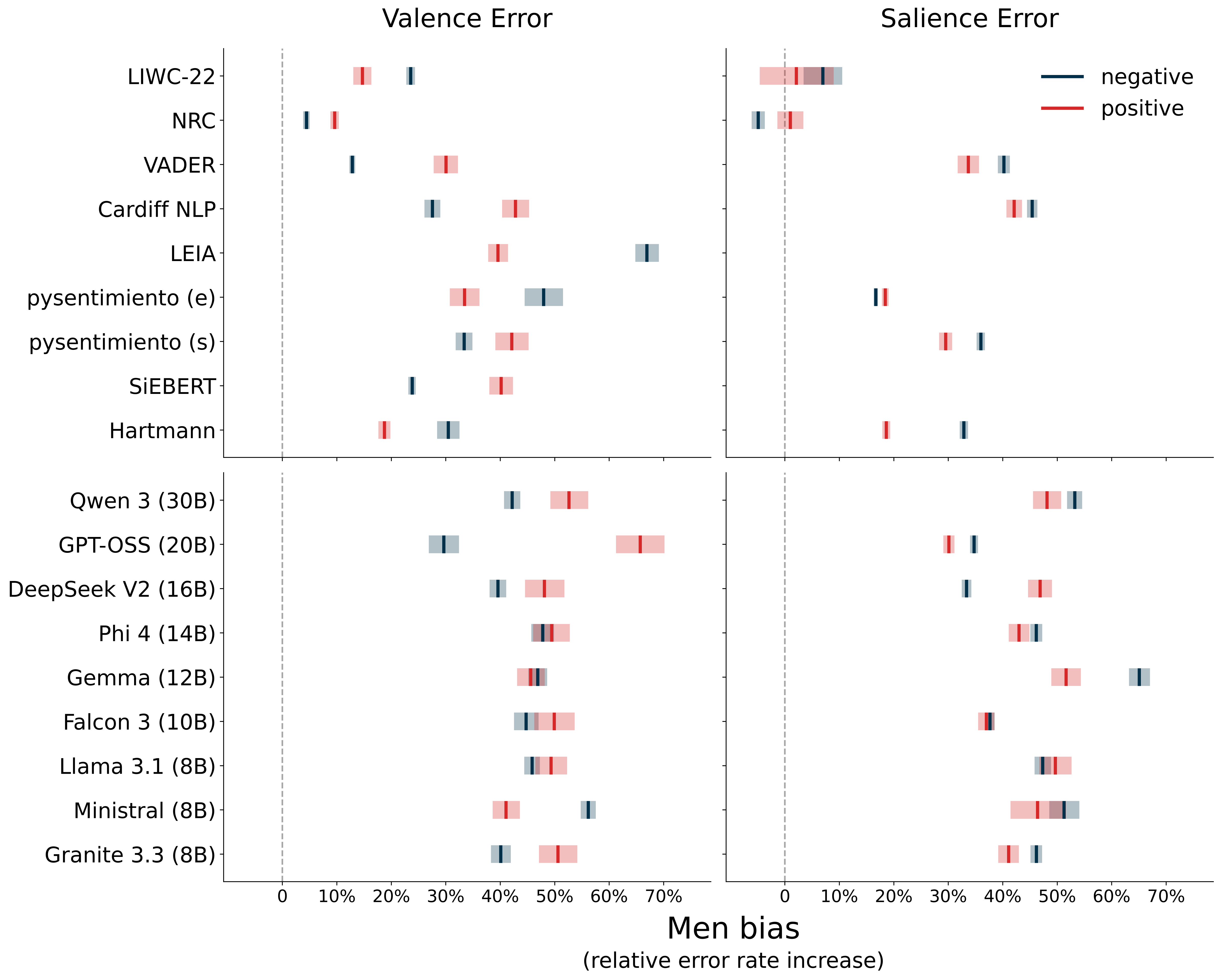}
\caption{\footnotesize{\textbf{Error rates for men are consistently higher than those for women across all model types for both positive and negative emotions.} The men bias is quantified as the relative increase in error rate, i.e. $(ER_{\text{men}} - ER_{\text{women}}) / ER_{\text{women}}$. Bootstrap resampling with $10,000$ iterations was used to compute 95\% confidence intervals. Note that while it might appear that the salience error for negative posts is significantly higher for women with the NRC-lexicon-based approach (negative value for the men bias), the corresponding p-value does not reach the significance threshold specified in the pre-registration.}}
\label{fig:bias}
\end{figure}

\begin{figure}[h!]
\centering
\includegraphics[width=14cm]{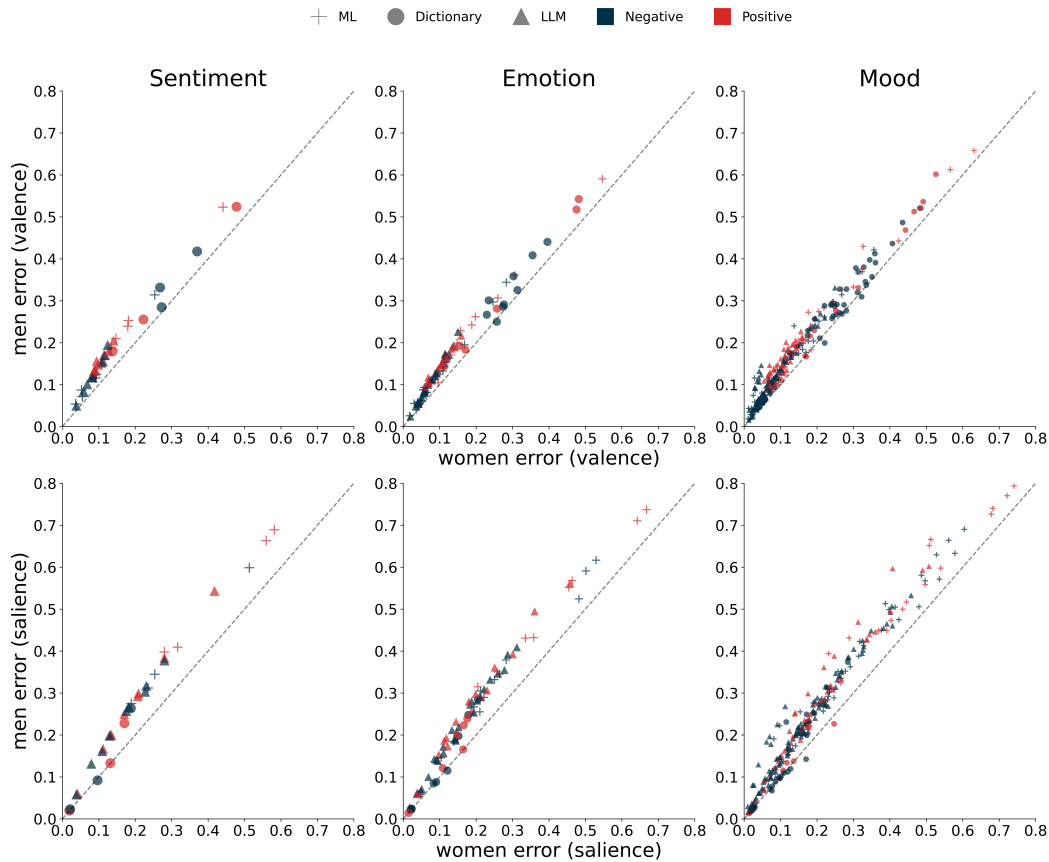}
\caption{\footnotesize{\textbf{Error rates for men are consistently higher than those for women across both types of error and all levels of analysis}: individual mood tags, emotion classes, and sentiment. The difference is more pronounced for salience error; however, only a limited number of models return neutral labels necessary for computing this error type.}}
\label{fig:levels}
\end{figure}

We find consistently higher error rates for men compared to women across all classifier types for both positive and negative emotions (\autoref{fig:bias}). This result cannot be explained by over-representation of specific emotions in our dataset as it is reproduced on a more granular level (\autoref{fig:levels}). In particular, according to pre-registered thresholds for $P$-values (note that Large Language Models have not been included in pre-registration and are not included in this summary):
\begin{itemize}
\item In all $18$ \textit{model} $\times$ \textit{sentiment} cases, valence errors for men are significantly higher ($P < 0.001$)
\item In $43$ of $45$ \textit{model} $\times$ \textit{emotion} cases, valence errors for men are significantly higher ($P < 0.001 / 5$)
\item In $115$ out of $144$ \textit{model} $\times$ \textit{mood tag} cases, valence errors for men are significantly higher ($P < 0.001 / 16$)
\end{itemize}
Notably, in no case are women's error rates significantly higher than men's. 
We find that the same pattern holds for classification done by LLMs with all tested models demonstrating higher error rate for men for both valence and salience errors (\autoref{fig:bias}).

\subsection*{Comparing to a case of human annotators}
Given the remarkable consistency of the results across different methods, there is reason to suspect that the source of bias might be in the labelling process, i.e., human-annotators might be worse at detecting men's emotional expressions from texts, which leads to incorrect labelling that, in turn, affects machine learning models.

To explore this hypothesis, we analysed the enISEAR dataset \citep{troiano2019crowdsourcing}, which allows us to compare self-reported emotions with third-party annotations while controlling for author gender. The dataset contains $1,001$ event descriptions provided by human authors with known gender and associated with a predefined emotion. For example, 'I felt \textit{disgust} when I read that hunters had killed one of the world-famous lions.' The emotion word is then removed, and annotators are asked to select the correct emotion from a predefined list ('anger', 'disgust', 'fear', 'guilt', 'joy', 'sadness', 'shame'). Each sentence is labeled by five annotators, resulting in $5,005$ labels in total.

Interestingly, the error rate for men is also larger in this dataset. The error rates are $27.8\%$ for men vs $23.6\%$ for women ($P = 0.001$) if all annotation labels are considered independently ($N = 5005$), and $23.4\%$ for men vs $18.2\%$ for women ($P = 0.047$) if each sentence is assigned a majority label ($N = 1001$).

\subsection*{Effects on downstream applications}

\begin{figure}[h!]
\centering
\includegraphics[width=8cm]{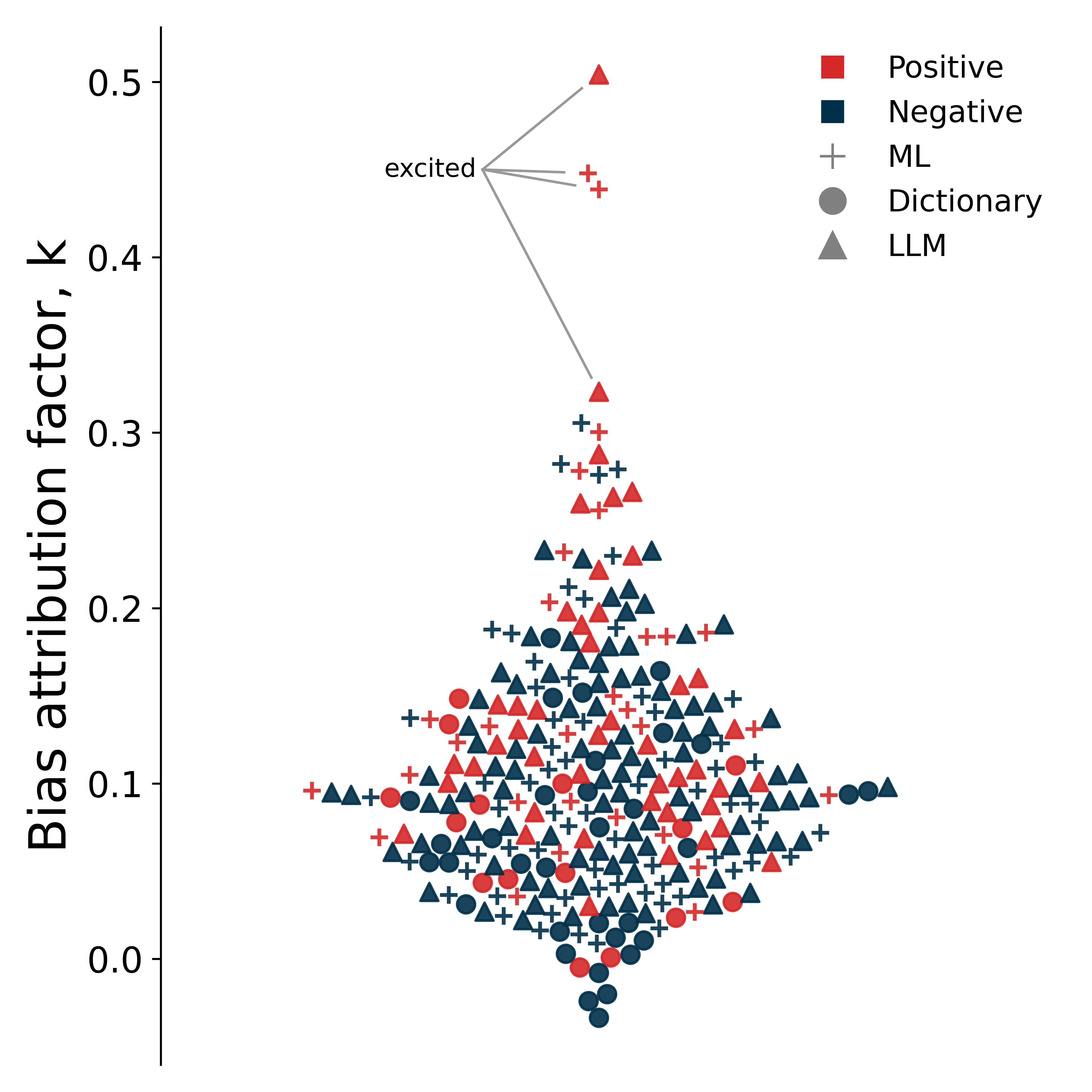}
\caption{\footnotesize{\textbf{Distribution of bias attribution factor across model-mood tag combinations shows that gender bias can substantially affect downstream applications.} The bias attribution factor $\hat{k}$ represents the percentage change in detected sentiment that can be attributed solely to gender composition differences between groups, assuming equal underlying emotional expression. The analysis shows that 48\% of model $\times$ mood tag combinations have bias attribution factors exceeding 10\%, which could meaningfully confound research conclusions.}}
\label{fig:effects}
\end{figure}

To understand if the observed gender bias is relevant in practice, we investigate whether it can meaningfully affect downstream applications of sentiment analysis. A common approach in such applications is to compute the proportion of posts expressing a certain emotion and compare it to a baseline. This approach can be used to compare two groups of people or to detect changes in sentiment over time.  For example, if the proportion of negative posts for one group is $f_1$ and for another group is $f_2$, where $f_1 > f_2$, then it is common to compute relative increase, $k = \frac{f_1 - f_2}{f_2} = \frac{f_1}{f_2} - 1$. For instance, if $f_1 = 30\%$ and $f_2 = 20\%$, then the reported increase is $k = 50\%$ as $\frac{0.3 - 0.2}{0.2} = 0.5$. For reference, events such as the election of the Prime Minister in Spain led to a $20\%$ increase in anxiety, the death of Kobe Bryant resulted in a $40\%$ increase in sadness in the US, and the Hanau shootings in Germany led to an $80\%$ increase in anger \citep{metzler2023collective}.

We estimate what values of $k$ can be explained not by differences in underlying emotions but by the bias we have discovered and the gender composition of the two groups. If the same fraction $f$ of people from two groups express a certain sentiment but the detection error rate for women is $e$ and for men is $e + \Delta$, then the following value of $k$ can be explained solely by gender composition of the two groups:

$$\hat{k} = \frac{f \times (1 - e)}{f \times (1 - (e + \Delta))} - 1 = \frac{1 - e}{1 - (e + \Delta)} - 1$$

We call this $\hat{k}$ the \textit{bias attribution factor} and use it as a measure of bias effect in downstream applications. We find that $\hat{k}$ might be as high as 50\% (\autoref{fig:effects}), and is larger than ten percent in $48\%$ of \textit{model} $\times$ \textit{mood tag} cases. We rank mood tags based on average $\hat{k}$ across all models (\autoref{tb:ranking}). This ranking could serve as a guide for which emotions researchers should be especially careful with. The average bias attribution factor is lowest for \textit{Annoyed}, \textit{Frustrated}, and \textit{Angry} mood tags, suggesting that these emotions are not much harder to detect in men. The highest values for the factor are for \textit{Excited}, \textit{Furious}, and \textit{Lonely}, indicating that automatic classifiers particularly struggle with detecting these emotions in men.

\section*{Discussion}

Our findings demonstrate that commonly used models underestimate sentiment and emotions expressed by men. These results were consistent across all tested approaches: carefully curated dictionaries, popular machine learning algorithms, and large language models. Our analysis was applied to a simple settings, where we counted an error only if classifier was returning a wrong sentiment, e.g. classifying a text with \textit{Anger} mood tag as positive. One might expect that when the task in more granular, e.g. distinguishing between two negative emotions such as \textit{Anger} and \textit{Fear} the errors and bias might be even higher.

Across two empirical settings, we observe a consistent gender-related disparity in emotion detection errors. In user-generated social media data, error rates are higher for texts authored by men. The same pattern appears in a controlled dataset constructed under experimental conditions in which the gender of the writer is recorded. In both settings, gender information is derived from observed metadata rather than inferred attributes, strengthening the ecological validity of the findings. The convergence of results across naturalistic and experimental data indicates that the observed gender disparity is not specific to a single dataset.
These results can be situated within the broader discourse on social bias in language technologies, which emphasises that systematic differences in model behaviour may arise from multiple factors, including data composition, annotation practices, and modelling choices \cite{navigli2023biases,hu2025generative,guilbeault2025age}. Our findings provide an empirical characterisation of gender-related error disparities in emotion detection and can serve as a basis for subsequent work aimed at examining underlying mechanisms and assessing implications for socially responsible language technologies.

The present findings further show that gender-related disparities in emotion detection accuracy can arise even under strictly gender-blind conditions: gender information is neither provided to annotators nor supplied to the model, yet error rates differ systematically by the gender of the writer. This result is notable in light of prior work on emotion-related text generation \cite{plaza-del-arco-etal-2024-angry}, which shows that when gender information is directly provided, large language models produce systematically different outputs for woman versus man personas. Taken together, these observations indicate that gender disparities can emerge both when demographic information is directly provided and when it is absent in the inputs to models. More broadly, the evaluation conditions in this study suggest that the observed gender-related differences represent a lower bound on performance disparities. This underscores the importance of examining model behaviour across demographic attributes, even in systems and evaluation settings that do not incorporate such information.

In practice, unequal emotion detection accuracy creates the potential for differential downstream consequences across demographic groups. Higher error rates for men may lead to systematic under-recognition of emotional signals, such as distress or negative affect, in applications where emotion detection informs subsequent decisions. This is particularly consequential in domains related to mental health or well-being support, where missed emotional signals may limit appropriate intervention. Conversely, lower error rates or higher sensitivity for women may result in more frequent attribution of emotional content in comparable settings. These asymmetries risk reinforcing existing associations between gender and emotion and may contribute to unequal downstream treatment across demographic groups.

For researcher, the immediate implication of our results is that researchers should account for potential gender biases when applying automatic classifiers to gender-heterogeneous samples, particularly when comparing groups with different gender compositions. Commonly measured changes in emotion levels on social media could be explained by these gender biases and thus results on a change in emotion frequency online might be explained simply by a change in activity levels across genders.

This study's findings point to several key limitations and promising directions for future research. First, while we identify a systematic bias, understanding the mechanisms behind it is a natural next step that could lead to more equitable sentiment classification models. Second, our analysis focuses on English language data. This provided a controlled context for assessing bias but may not capture the full diversity of emotional expression across linguistic or cultural settings. Extending this study to other languages would therefore be valuable for testing the robustness of our findings and enabling comparative analyses of error profiles across contexts.

Third, our analysis of gender was limited to a binary construct. This simplification overlooks potential heterogeneity within and across gender categories. Consequently, understanding variations within demographic groups, rather than only between them, remains an important challenge for future work. Addressing this challenge will require data sources that better reflect the complexity of gender identity.

Fourth, ethical, practical, and legal constraints shaped the set of models we evaluated. To ensure user privacy, we refrained from using third-party services, which limited our inclusion of larger open-weight or commercial models. Consequently, the nature and magnitude of bias in these larger models remain an open question. Finally, our focus on zero-shot prompting established a necessary first step; analysing alternative prompting strategies is thus an interesting future direction.

This study lays the groundwork for a more equitable examination of fairness in affective text analysis. Our findings call for a renewed collective effort to ensure that this field advances responsibly. Researchers and practitioners share the responsibility to recognise and address this challenge, working toward future systems that can more accurately and equitably identify and interpret emotional expressions across diverse user populations.

\section*{Methods}\label{sec:methods}
The data was collected from TalkLife, a social media platform specifically designed for users to share their emotional experiences. Users' posts were gathered in accordance with TalkLife's data usage policies. The dataset was provided under a licensing agreement with TalkLife. The dataset comprises $6,633,562$ posts made by $316,387$ users from 19/08/2011 to 14/01/2019. The data was analysed following the pre-registered research design.

Each post on the platform is tagged with a mood label that authors use to communicate their emotional state. Overall, there are $64$ mood tags in the dataset, ranging from \textit{Afraid} to \textit{Worried}. Since some of these categories may be too nuanced for a text classification task, we selected $16$ tags and mapped them to $5$ broader categories based on dimensional models of emotion from affective science literature, as suggested by \cite{aroyehun2023leia}. Following the pre-registered design, we conducted and reported analysis on three levels: 1) for each of the $16$ TalkLife mood tags separately, 2) for each of the $5$ emotional state categories (\textit{Sadness}, \textit{Anger}, \textit{Fear}, \textit{Affection}, and \textit{Happiness}), and 3) by combining these $5$ categories into a binary sentiment label of \textit{Positive} or \textit{Negative}. The list of selected mood tags and their corresponding mappings is presented in \autoref{tb:map}. TalkLife users can select from \textit{Nonbinary}, \textit{Female}, and \textit{Male} genders, or provide their own value for this field. For our analysis, we focused on the \textit{Female} and \textit{Male} categories as other categories do not have sufficient data for statistical analysis. There are $1,738,282$ posts with one of the sixteen mood tags from $147,980$ users with known binary gender.

It could be meaningless to apply sentiment or emotion classifiers to texts that are too short. Additionally, some machine learning classifiers have an upper limit on the size of texts that can be used as input. For that reason, we filtered out posts that are less than $5$ or more than $512$ tokens as defined by the RoBERTa tokenizer \citep{liu2019roberta} with these thresholds defined in our pre-registration. There were $26,768$ posts not meeting these criteria, resulting in a final dataset of $1,711,514$ posts from $146,883$ users. The distribution of posts across mood tags is displayed in \autoref{fig:distribution}.

We then applied commonly used machine learning classifiers, popular dictionary-based methods, and large language models (LLMs) (see \autoref{tb:model_list} for the full list of models) to the posts. The same prompt was used for all LLMs (\autoref{tb:prompt}). To preserve users privacy, we only use open-weight LLMs and run them locally. Due to high computational costs and the fast-changing landscape of LLMs, this part of the analysis was not included in the pre-registration, and, thus, the results for LLMs should be treated as exploratory.

Since different models use different output labels (e.g., NRC includes the label `trust' while other models do not), we standardised all labels to one of three sentiment categories: \textit{positive}, \textit{negative}, or \textit{neutral} using the mapping provided in \autoref{tb:mapping}. We then computed two error values for each mood tag: the \textit{valence error} as the proportion of cases where classifiers return positive labels instead of negative self-labels, or vice versa; and the \textit{salience error} as the proportion of neutral labels returned by classifiers instead of negative or positive self-labels. Note that \textit{salience error} can be computed only for those classifiers that could return a \textit{neutral} label. We have checked for the significance of difference between women and men error rates using $\chi^2$-test with Bonferroni-corrected thresholds as defined in pre-registration.

For computing effect size, we calculated total error rather than valence and salience errors, as this better reflects real-world settings. Please note that total error is not equal to the sum of valence and salience errors. For example, if the actual sentiment is negative but it is predicted to be positive in $33.3\%$ of cases and neutral in $33.3\%$ of cases, then according to our definition, the valence error is $50\%$, the salience error is $50\%$, and the total error is $66.6\%$.

\section*{Data and Code}
The dataset is licensed from TalkLife. The dataset can be accessed by contacting TalkLife directly for information on licensing terms and availability.

We publicly share model predictions without actual post content at The Open Science Framework: \url{https://doi.org/10.17605/OSF.IO/AP24D}.

The code used to computer the results is available at \url{https://doi.org/10.17605/OSF.IO/AP24D}.

\bibliographystyle{unsrtnat}
\bibliography{refs}

\clearpage
\appendix
\renewcommand{\thesection}{S\arabic{section}}
\renewcommand{\thesubsection}{S\arabic{section}.\arabic{subsection}}

\setcounter{figure}{0}
\renewcommand{\thefigure}{S\arabic{figure}}
\renewcommand{\figurename}{Figure}

\setcounter{table}{0}
\renewcommand{\thetable}{S\arabic{table}}
\renewcommand{\tablename}{Table}

\section*{Supplementary Materials}

\begin{figure}[htp]
\centering
\includegraphics[width=10cm]{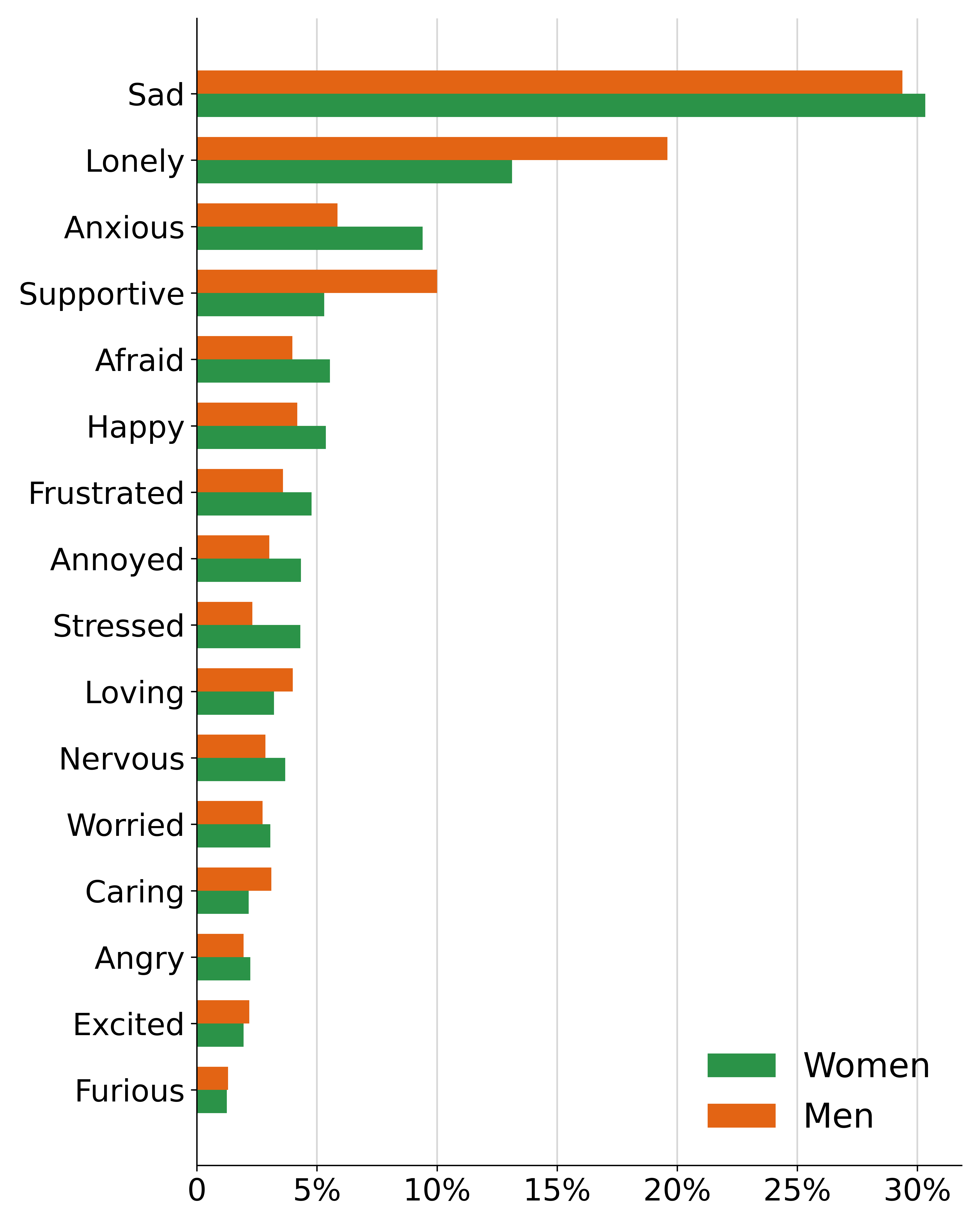}
\caption{\footnotesize{\textbf{Distribution of posts across mood tags.} The figure displays the proportion of posts labeled with a certain mood tag, where proportions are calculated within gender groups. Mood tags are ordered by their overall prevalence across all posts in the dataset.}}
\label{fig:distribution}
\end{figure}

\begin{table}[ht]
\centering
\caption{Mapping of TalkLife mood tags to emotion labels}
\begin{tabular}{|l|l|p{10cm}|}
\hline
\textbf{Sentiment} & \textbf{Emotion} & \textbf{TalkLife mood tags} \\
\hline
\multirow{3}{*}{Negative} & Sadness & Sad, Lonely \\
\cline{2-3}
& Anger & Angry, Annoyed, Frustrated, Furious \\
\cline{2-3}
& Fear & Anxious, Stressed, Afraid, Nervous, Worried \\
\hline
\multirow{2}{*}{Positive} & Affection & Loving, Caring, Supportive \\
\cline{2-3}
& Happiness & Happy, Excited \\
\hline
\end{tabular}
\label{tb:map}
\end{table}

\begin{table}[h]
\centering
\caption{List of sentiment analysis models and methods evaluated in this study}
\begin{tabular}{p{0.95\linewidth}}
\toprule
\textbf{Transformer-based Classifiers} \\
LEIA: Linguistic Embeddings for the Identification of Affect \citep{aroyehun2023leia} \\
pysentimiento: A Python toolkit for Sentiment Analysis and Social NLP tasks \citep{perez2021pysentimiento} \\
Emotion English DistilRoBERTa-base \citep{hartmann2022emotionenglish} \\
Twitter-roBERTa-base for Sentiment Analysis \citep{camacho-collados-etal-2022-tweetnlp} \\
SiEBERT - English-Language Sentiment Classification \citep{hartmann2023} \\[2ex]

\textbf{Dictionary-based methods} \\
VADER: Valence Aware Dictionary and sEntiment Reasoner \citep{hutto2014vader} \\
NRC Word-Emotion Association Lexicon \citep{mohammad2013crowdsourcing} \\
LIWC-22: Linguistic Inquiry and Word Count \citep{boyd2022development} \\[2ex]

\textbf{Large Language Models} \\

DeepSeek V2 Lite Chat (16B) \citep{deepseekv2} \\
Falcon 3 (10B) \citep{Falcon3} \\
Gemma 3 (12B) \citep{gemma_2025} \\
GPT-OSS (20B) \citep{openai2025gptoss120bgptoss20bmodel} \\
Granite 3.3 (8B) \citep{granite2025} \\
Llama 3.1 (8B) \citep{meta2024} \\
Ministral (8B) \citep{ministral2024} \\
Phi 4 (14B) \citep{phi2024} \\
Qwen 3 (30B) \citep{qwen3technicalreport} \\
\bottomrule
\end{tabular}
\label{tb:model_list}
\end{table}

\begin{table}[h]
\centering
\caption{Prompt used for sentiment classification by LLMs}
\begin{tabular}{|l|p{0.8\linewidth}|}
\hline
\textbf{Role} & \textbf{Content} \\
\hline
system & Please perform Sentiment Classification task. Given the sentence, assign a sentiment label from ['negative', 'neutral', 'positive']. Return your response in JSON format with the key 'sentiment' and the label as the value. \\
\hline
user & \{\{CONTENT OF POST\}\} \\
\hline
assistant & \{"sentiment": "\textcolor{blue}{\{\{\{MODEL OUTPUT START HERE\}\}\}} \\
\hline
\end{tabular}
\label{tb:prompt}
\end{table}

\begin{table}[ht]
\centering
\caption{Mapping of model outputs to one of the three sentiment labels}
\begin{tabular}{|p{8cm}|p{8cm}|}
\hline
\textbf{Model Name} & \textbf{Mapping} \\
\hline
LEIA: Linguistic Embeddings for the Identification of Affect  & 
Same as in \autoref{tb:map} \\[2cm]
pysentimiento: A Python toolkit for Sentiment Analysis and Social NLP tasks (sentiment model) & 
As is (positive, negative, neutral) \\[2cm]
pysentimiento: A Python toolkit for Sentiment Analysis and Social NLP tasks (emotion model) & 
joy\kern3pt $\rightarrow$\kern3pt positive, (sadness$\|$anger$\|$disgust$\|$fear)\kern3pt $\rightarrow$\kern3pt negative, (surprise$\|$others)\kern3pt $\rightarrow$\kern3pt neutral \\[2cm]
Emotion English DistilRoBERTa-base &
joy\kern3pt $\rightarrow$\kern3pt positive, (sadness$\|$anger$\|$disgust$\|$fear)\kern3pt $\rightarrow$\kern3pt negative, (surprise$\|$neutral)\kern3pt $\rightarrow$\kern3pt neutral \\[2cm]
Twitter-roBERTa-base for Sentiment Analysis &
As is (positive, negative, neutral) \\[2cm]
SiEBERT - English-Language Sentiment Classification &
As is (positive, negative) \\[1cm]
\hline
VADER: Valence Aware Dictionary and sEntiment Reasoner &
(compound score $\ge$ 0.05)\kern3pt $\rightarrow$\kern3pt positive,
(compound score $\le$ $-0.05$)\kern3pt $\rightarrow$\kern3pt negative,
($-0.05 <$ compound score $< 0.05$)\kern3pt $\rightarrow$\kern3pt neutral\\[2cm]
NRC Word-Emotion Association Lexicon &
(joy$\|$trust)\kern3pt $\rightarrow$\kern3pt positive, (sadness$\|$anger$\|$disgust$\|$fear)\kern3pt $\rightarrow$\kern3pt negative, (surprise$\|$anticipation)\kern3pt $\rightarrow$\kern3pt neutral\\[2cm]
LIWC-22: Linguistic Inquiry and Word Count &
(Tone $\ge$ 55)\kern3pt $\rightarrow$\kern3pt positive,
(Tone $\le$ 45)\kern3pt $\rightarrow$\kern3pt negative,
($45 <$ Tone $< 55$)\kern3pt $\rightarrow$\kern3pt neutral\\[1cm]
\hline
\end{tabular}
\label{tb:mapping}
\end{table}

\begin{table}[ht]
\centering
\caption{Average bias attribution factor for different mood tags}

\begin{tabular}{|p{4cm}|p{4cm}|}
\hline
\textbf{Mood tag} & \textbf{Average $\hat{k}$} \\
\hline
Excited & 0.237 \\
Furious & 0.166 \\
Lonely & 0.155 \\
Supportive & 0.138 \\
Caring & 0.133 \\
Anxious & 0.124  \\
Nervous & 0.121 \\
Worried & 0.117 \\
Happy & 0.097 \\
Afraid & 0.091 \\
Sad & 0.086 \\
Stressed & 0.085 \\
Loving & 0.074 \\
Angry & 0.053 \\
Frustrated & 0.041 \\
Annoyed & 0.038 \\
\hline
\end{tabular}
\label{tb:ranking}
\end{table}

\end{document}